\documentclass[letterpaper,10pt,conference]{ieeeconf}

\IEEEoverridecommandlockouts%
\makeatletter
\def\IEEElabelanchoreqn#1{\bgroup%
\def\@currentlabel{\p@equation\theequation}\relax
\def\@currentHref{\@IEEEtheHrefequation}\label{#1}\relax
\Hy@raisedlink{\hyper@anchorstart{\@currentHref}}\relax
\Hy@raisedlink{\hyper@anchorend}\egroup}
\makeatother
\newcommand{\subnumberinglabel}[1]{\IEEEyesnumber\IEEEyessubnumber*\IEEElabelanchoreqn{#1}}

\makeatletter
\let\NAT@parse\undefined%
\makeatother

\usepackage{mathtools,amssymb}
\DeclarePairedDelimiterX{\norm}[1]{\lVert}{\rVert}{#1}
\DeclarePairedDelimiterX\set[1]\lbrace\rbrace{#1}
\usepackage[usenames,dvipsnames]{xcolor}
\usepackage{graphicx}
\usepackage[linesnumbered,ruled,vlined]{algorithm2e}
\usepackage[pdfusetitle,colorlinks,unicode,breaklinks]{hyperref}
\hypersetup{
	pdfcreator  = {Iordanis Chatzinikolaidis},
    pdfsubject  = {Robotics},
    pdfkeywords = {Legged Robots, Trajectory Optimization, Optimization, Optimal Control, Simulation, Dynamic Programming},
    breaklinks  = true,
    colorlinks  = true,
	urlcolor    = Black,
	linkcolor   = Black,
	citecolor   = Black
}
\usepackage[all]{hypcap} % Links at the top of floats
\usepackage[capitalise,nameinlink]{cleveref}
\crefname{section}{Sec.}{Secs.}
\crefname{algorithm}{Algo.}{Algos.}
\crefname{appsec}{Appendix}{Appendices}
\crefname{equation}{}{}
\usepackage{xspace}% Common abbreviations
\newcommand{\eg}{\hbox{\emph{e.g.}}\xspace}
\newcommand{\ie}{\hbox{\emph{i.e.}}\xspace}
\newcommand{\etc}{\hbox{\emph{etc.}}\xspace}
\usepackage{import}
\usepackage{booktabs}
\SetKwInput{KwInput}{Input}
\SetKwInput{KwOutput}{Output}
\let\labelindent\relax
\usepackage{enumitem}
\setitemize{leftmargin=*}
\usepackage[style=ieee,maxnames=1,minnames=1]{biblatex}
\AtBeginBibliography{\footnotesize}

\usepackage{tikz}
\newcommand\copyrighttext{This is a preprint / authors' own draft version of paper prior to peer-reviewed publication.}
\newcommand\copyrightnotice{%
\begin{tikzpicture}[remember picture,overlay]
  \node[anchor=south,yshift=30pt] at (current page.south) {\fbox{\parbox{\dimexpr0.73\textwidth-\fboxsep-\fboxrule\relax}{\copyrighttext}}};
\end{tikzpicture}%
}

\title{\LARGE \bf
  Trajectory Optimization for Contact-rich Motions using \\ Implicit Differential Dynamic Programming
}
\author{Iordanis Chatzinikolaidis and Zhibin Li%
\thanks{This research is supported by the EPSRC as part of the CDT in Robotics and Autonomous Systems (EP/L016834/1) and the EPSRC UK RAI Hub in Offshore Robotics for Certification of Assets (ORCA) (EP/R026173/1).}
\thanks{Authors are with the School of Informatics and the Edinburgh Centre for Robotics, University of Edinburgh, Edinburgh, {UK}. Emails: \texttt{\small iordanis.cs@gmail.com}; \texttt{\small zhibin.li@ed.ac.uk}}%
}

\begin{document}

\maketitle
\copyrightnotice%
\thispagestyle{empty}
\pagestyle{empty}

%%%%%%%%%%%%%%%%%%%%%%%%%%%%%%%%%%%%%%%%%%%%%%%%%%%%%%%%%%%%%%%%%%%%%%%%%%%%%%%%
\begin{abstract}
This paper presents a novel approach using sensitivity analysis for generalizing Differential Dynamic Programming (DDP) to systems characterized by implicit dynamics, such as those modelled via inverse dynamics and variational or implicit integrators.
It leads to a more general formulation of DDP, enabling for example the use of the faster recursive Newton-Euler inverse dynamics.
We leverage the implicit formulation for precise and exact contact modelling in DDP, where we focus on two contributions: (1) Contact dynamics in acceleration level that enables high-order integration schemes; (2) Formulation using an invertible contact model in the forward pass and a closed-form solution in the backward pass to improve the numerical resolution of contacts.
The performance of the proposed framework is validated (1) by comparing implicit versus explicit DDP for the swing-up of a double pendulum, and (2) by planning motions for two tasks using a single leg model making multi-body contacts with the environment: standing up from ground, where a priori contact enumeration is challenging, and maintaining balance under an external perturbation.
\end{abstract}
%%%%%%%%%%%%%%%%%%%%%%%%%%%%%%%%%%%%%%%%%%%%%%%%%%%%%%%%%%%%%%%%%%%%%%%%%%%%%%%%

\section{Introduction}

The long-standing research goal of creating robots capable of physically interacting with our environment remains elusive.
Typical tasks, such as moving around the environment (locomotion) and modifying the surroundings (manipulation), ultimately require a complex sequence of physical contacts between the robot and the external world.
Achieving such capabilities requires effective solutions for producing contact-rich motions.

To date, we still have limited technologies to replicate animal- or human-level interaction skills on robots.
This observation forces us to rethink the root of these limitations, which is more at an algorithmic and theoretical level rather than in hardware; it is nowadays possible yet difficult to validate a large range of physical capabilities in high-fidelity physics simulation.
The scope here is on producing contact-rich motions for robot locomotion, leaving the applicability and adaptation on robot manipulation for future work.

\begin{figure}[t]
	\centering
	\includegraphics[width=0.24\linewidth]{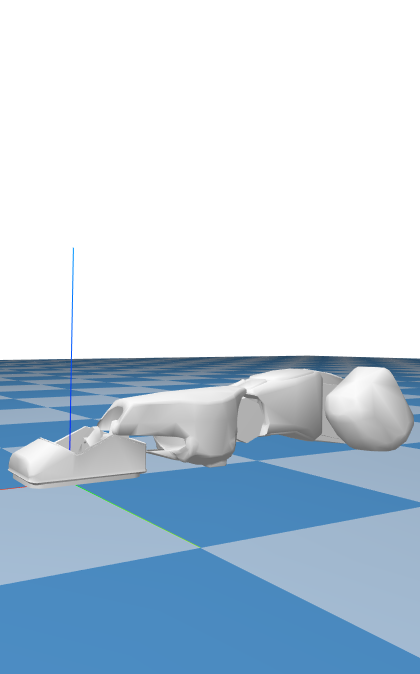}
	\includegraphics[width=0.24\linewidth]{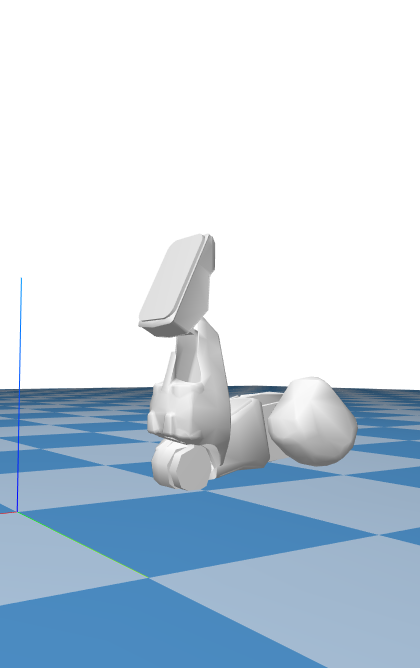}
	\includegraphics[width=0.24\linewidth]{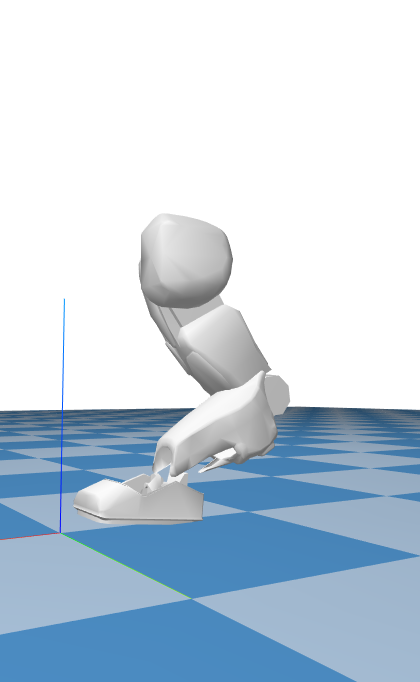}
	\includegraphics[width=0.24\linewidth]{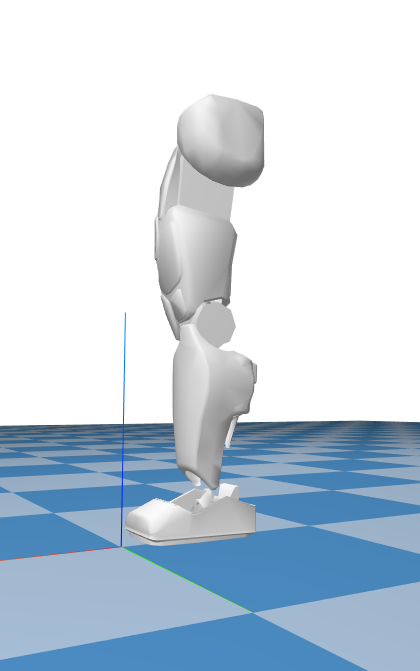}
	\includegraphics[width=\linewidth]{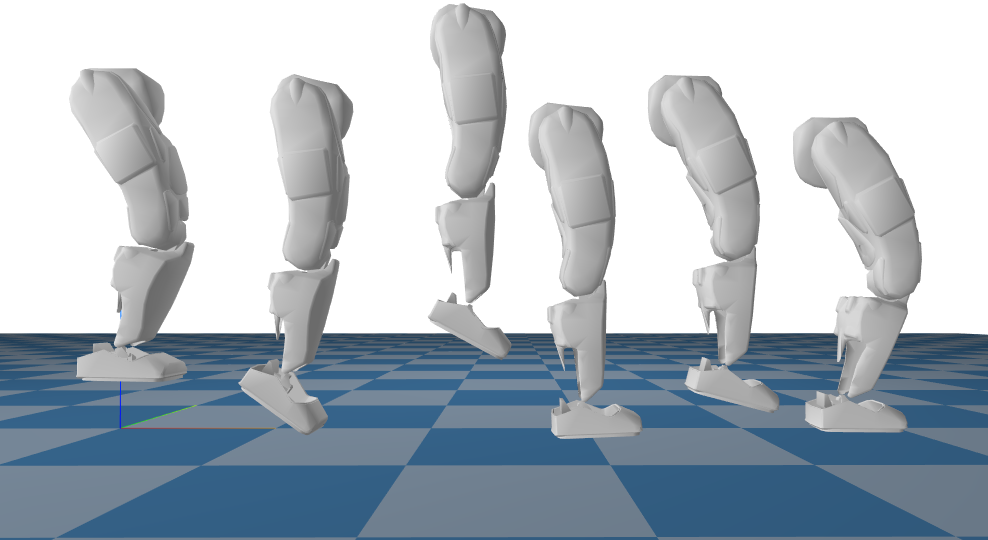}
	\caption{Complex multi-contact motions of a single leg robot computed by the proposed framework in time-lapsed snapshots: dynamic standing up from the ground (top), and balancing against an external perturbation (bottom).}%
	\label{fig:teaser}
	\vspace{-5mm}
\end{figure}

\subsection{Machine Learning}

Machine learning techniques can be integrated with optimization, where cost-preference learning alleviates the burden of manual design of cost functions for optimization-based planning and control.
This achieved rough-terrain quadrupedal locomotion for Little-Dog robot~\cite{zucker2011}.
Going beyond such a hierarchy of complex integration of learning, footstep planning, trajectory optimization and reflex control, reinforcement learning for continuous control provides an alternative in an end-to-end fashion.
A reinforcement learning approach trained by an adaptive terrain curriculum demonstrated robust single-skill trotting that traversed a variety of indoor and outdoor unstructured environments~\cite{lee2020}.
An architecture of multi-expert reinforcement learning is able to extend capabilities to multi-skill and multi-modal locomotion with coherent fall recovery, trotting, and all dynamic transitions in-between different modes~\cite{yang2020}.
These neural network based feedback policies were trained in simulation and then deployed on real robots, but still demonstrated robustness under scenarios that were never encountered during training. 

However, these neural network policies act as reactive feedback control that responds to the proprioceptive state feedback.
It is hard to incorporate future target objectives for long-term temporal planning.
Even though they are computationally fast to run in real time, it is difficult to guarantee the long-term stability and optimality of motions, \ie~whether the robot will fall in the future or whether a successful sequence of motions is more optimal in terms of energy efficiency and sufficient stability margins against uncertainties.
Moreover, for safety critical applications, such an approach is not able to provide verifiable validity before execution.
To overcome these limitations, a more mathematically principled approach enables to take into account knowledge about the constraints of the robot and environments, and to provide verifiable long-term stability and feasibility.

\subsection{Model-based Mathematical Optimization}

Trajectory optimization (TO) has attracted increasing research interest for motion planning and control of highly dynamical, underactuated robots~\cite{kelly2017,xin2020}.
Similar to learning approaches, it also has the potential of generating complex motions in a high-level manner:
A user can design and specify a desired task using physical terms with associated weights via a cost function~\cite{todorov2018}, which can also be automatically tuned~\cite{yuan2019}.
This approach is quite flexible, encompassing a wide-range of cases; for example, additional contact points can be included in the optimization to increase the robustness against perturbations for loco-manipulation tasks~\cite{wolfslag2020}.

This is particularly interesting for robotic systems that need to compute through-contact motion plans, which involve multiple contact interactions.
Physical contacts are traditionally difficult to model and incorporate in motion planning frameworks.
Most approaches are hybrid, in the sense that the contact schedule pattern~\cite{chatzinikolaidis2018,wang2020} or corresponding timings are provided a priori, while contacts are desirable with the end-effectors only.
This leads to difficulties in practical implementations because selection of locations and timings is in general non-trivial, while restricting contacts to end-effectors only limits the motion repertoire.

DDP---a prominent shooting TO methodology~\cite{mayne1966}---is among the most promising approaches in terms of efficiency for through-contact motion planning.
This is demonstrated by a multitude of previous works that used DDP as backbone:
From impressive results in simulation~\cite{tassa2012}, to real-time applications for high-dimensional legged robots~\cite{neunert2017a,carius2019}.
However, properly modelling contacts has proven a considerable challenge; most DDP implementations resort to approximations and simplifications that require well-tuned contact parameters.
A fundamental reason is that contact phenomena are canonically described implicitly.

The original DDP algorithm and its subsequent studies assume that the discrete-time systems considered are explicitly defined.
While this is valid in broad terms, it excludes implicitly defined dynamical systems~\cite{hairer1993}.
These are typically more challenging because they require the solution of nonlinear equations.
However, they offers computational advantages, \eg~providing stability even for stiff differential equations.
Further, handling implicitly defined systems allows more principled contact modelling in DDP.

\subsection{Contributions}

In this work, we focus on the optimization paradigm and provide \textit{theoretical} and \textit{algorithmic} contributions as:
\begin{itemize}
	\item An extension to the DDP algorithm that handles explicitly and implicitly defined systems in a unified manner.

	\item Based on this, we propose an approach leveraging an invertible model~\cite{todorov2014} for exact contact resolution in DDP.

	\item Results demonstrating the possibility of exploiting properties of implicit integrators in DDP settings.
\end{itemize}

We benchmark our extension by applying it on implicitly and explicitly defined models, and on two cases of multi-contact whole-body motion planning for a planar single leg robot that makes multi-body contacts:
Standing-up from ground and balancing from an initial perturbation in a receding horizon fashion (\cref{fig:teaser}).
Our approach is equally applicable to models with large degrees of freedom and arbitrary contact configurations, such as using multiple legs.

The remaining sections are organized as:
\cref{sec:prior_work} discusses prior work and extensions of the DDP algorithm, and applications of DDP for through-contact motion planning.
\cref{sec:preliminaries} summarizes DDP and how contacts are typically resolved in simulation.
In \cref{sec:implicit_DDP}, we present our extension of DDP and, in \cref{sec:acceleration_level}, how to utilize it for through-contact planning.
\cref{sec:results} provides comparisons between explicit and implicit systems in the context of DDP, and two motion planning studies for a single leg standing up and balancing in multi-contact settings.
We summarize and conclude in \cref{sec:conclusion}.

\section{Prior Work on Differential Dynamic Programming}%
\label{sec:prior_work}

\subsection{Differential Dynamic Programming}

DDP was originally introduced in~\cite{mayne1966}.
Its main advantage with respect to the Dynamic Programming algorithm~\cite{bellman1962} is that it does not suffer from the curse of dimensionality by sacrificing global optimality.
Subsequently, a number of improvements and extensions of DDP have been introduced.
Recently, there was a resurgence of interest due to its potential for efficient planning for high-dimensional systems.

DDP is a second-order algorithm that exhibits quadratic convergence similar to Newton's methods~\cite{liao1992}.
Thus, it requires second-order information, which can be computationally challenging for high-dimensional models.
To resolve this, the iLQR variant performs a Gauss-Newton approximation of the Hessian based on first-order information only, albeit with superlinear convergence~\cite{todorov2005}.

The original DDP algorithm is concerned with unconstrained discrete dynamical systems only.
Control bounds can be considered via a projected Newton quadratic programming (QP) solver~\cite{tassa2014}.
More general nonlinear inequality constraints via an active-set method~\cite{xie2017}.
In robotics, it is common to consider multiple tasks in a hierarchical fashion, which is possible to do for DDP too~\cite{geisert2017}.
In legged locomotion, the discontinuous nature of contact phenomena has led to the development of tailored approaches.
For example, a predefined gait pattern and centroidal dynamics model was considered in~\cite{budhiraja2018}, and more general hybrid systems in~\cite{li2020}.
We underline that the DDP framework presented next can incorporate the previous extensions straightforwardly.

Finally, a brief discussion about the application of DDP for implicitly defined systems from a Lie theoretic viewpoint is given in~\cite{boutselis2020}.
Here, we present a more complete and deep treatment, with extensive comparisons.
Furthermore, our vector-based formulation is much more familiar and common for robotic systems applications.

\subsection{Through-contact Motion Planning}

Applications of DDP to motion planning and control for legged robots have been very impressive.
From simple, approximate models up until whole-body models, DDP provides a means for fast and even real-time computations.

In~\cite{tassa2012}, DDP was used to control a humanoid model.
A diverse set of behaviours was generated by simply changing weights in the cost function through a GUI interface.
An approximate solution for the contact dynamics was used, with a contact model similar to the one that is used here.
In this work, the implicit formulation that we present next allows the consideration of contacts in DDP without requiring approximations to the contact model itself.

For quadruped robots, a diverse set of motions both in simulation and in hardware was shown in~\cite{neunert2017a}.
To take into account contacts, a nonlinear spring-damper model was used.
Even though tuning for each contact is done independently, spring-damper models can be difficult to tune in practice and require very small time steps.
It is common for the optimizer to explore states where the current model parameters are not valid, while the small time steps translate into a large problem.
Here, in the forward pass the model takes into account all possible contacts in a centralized manner (through the coupling with the contact-space inertia matrix), while independently solve for each contact at the backward pass (by leveraging our implicit DDP formulation and the model's invertibility).
Thus, performance is similar to complementarity formulations with large time steps, while we are capable to compute straightforwardly gradients in the backward pass.

To eliminate the unrealistic effects of spring-damper models, a hard contact model was used in~\cite{carius2018}.
Unfortunately, contact impulses require the numerical solution of a quadratically constrained quadratic program (QCQP), typical in time-stepping approaches with unilateral and friction cone constraints, and formulates the problem in a bilevel fashion.
This complicates the derivative computation due to the numerical nature of the solution.
We resolve this issue by leveraging the invertibility of the contact model: in the forward pass, the QCQP is solved with the associated constraints; in the backward pass, a closed-form computation is used that avoids the bilevel formulation.
As a result, this does not pose issues with differentiation and leads to a faster and simpler implementation, without the need for backpropagation.

A multiple shooting variant was presented in~\cite{mastalli2020}, extending the work in~\cite{giftthaler2018a}.
It allows easier initialization since both state and control sequences can be used.
Unfortunately, the intermediate iterates of the algorithm are infeasible, meaning that early stopping with a feasible trajectory, as in DDP, is not possible.
This is a necessary property in our case, since the through-contact motion planning approach that we present is running in a receding horizon fashion.
Furthermore, the contact schedule is predefined in~\cite{mastalli2020}, while here contacts are activated according to the natural dynamics of the system~\cite{posa2014}.
Finally, friction cone constraints are neglected or can be taken into account through penalization in the cost function, which can be in practice difficult to tune and can lead to unrealistic solutions.
Due to the imposition of contacts as equality constraints, attractive forces can arise at the solution, violating the unilateral constraint.
Our framework here utilizes full unilateral and friction cone contact constraints without any approximation or penalization.

\section{Preliminaries}%
\label{sec:preliminaries}

\subsection{Summary of Differential Dynamic Programming}

DDP is concerned with the optimization of a performance criterion for an unconstrained discrete-time dynamical system~\cite{mayne1966,tassa2012}.
This can be expressed as
\begin{IEEEeqnarray*}{r'l}%
	\subnumberinglabel{eq:general_opt}
    \underset{u_i}{\text{min}} & l_f(x_N) + \sum\nolimits_{i=t}^{N-1} l_i(x_i, u_i)\\
    \text{s.t.} & x' = f(x, u).\label{eq:explicit_dynamics}
\end{IEEEeqnarray*}
Here, \(l_i\) is an additive cost at time step \(i\) and \(l_f\) is the final cost, \(x_i\) and \(u_i\) are the state and control, \(N\) is the length of the horizon, while \(\cdot'\) denotes the quantity at the next time step, \eg~the next state in our context.

According to the principle of optimality,~\eqref{eq:general_opt} can be expressed via the value function, which is the total cost at a given state once we apply the optimal control sequence.
The principle of optimality makes the computation of the value function iterative, and at a state \(x\) is given by
\begin{equation*}
    V(x) = \min_u l (x,u) + V' (x') = \min_u l (x,u) + V' (f (x,u)).
\end{equation*}
Since finding the global minimum is challenging, DDP performs a quadratic approximation of the value function and subsequently improves the control sequence \(\{u_i\}\) locally.
If we define the \(Q\)-function as
\begin{equation}%
	\label{eq:Q_fun}
    Q(x,u) = l(x,u) + V'(x'),
\end{equation}
a quadratic approximation about the current point \((x_i, u_i)\) is
\begin{IEEEeqnarray}{rCl}%
	\label{eq:Q_fun_quad_expansion}
    Q(x, u) & \approx & Q(x_i,u_i) + Q_x(x_i,u_i) \delta x + Q_u(x_i,u_i) \delta u\nonumber\\
    && +\> \tfrac{1}{2} \begin{bmatrix} \delta x \\ \delta u \end{bmatrix}^T \begin{bmatrix} Q_{xx}(x_i,u_i) & Q_{xu}(x_i,u_i) \\ Q_{ux}(x_i,u_i) & Q_{uu}(x_i,u_i) \end{bmatrix} \begin{bmatrix} \delta x \\ \delta u \end{bmatrix} \ \ \
\end{IEEEeqnarray}
while \(\delta x = x - x_i\) and \(\delta u = u - u_i\) are state and input perturbations.
The terms in~\eqref{eq:Q_fun_quad_expansion} are computed by expanding and matching same terms in \cref{eq:Q_fun} as
\begin{equation}%
	\label{eq:Q_fun_quad_terms}
    \begin{IEEEeqnarraybox}[\IEEEeqnarraystrutmode\IEEEeqnarraystrutsizeadd{2pt}{2pt}][c]{rCl}
        Q_x & = & l_x + V'_{x'} f_x\\
        Q_u & = & l_u + V'_{x'} f_u  
    \end{IEEEeqnarraybox} \quad
    \begin{IEEEeqnarraybox}[\IEEEeqnarraystrutmode\IEEEeqnarraystrutsizeadd{2pt}{2pt}][c]{rCl}
        Q_{xx} & = & l_{xx} + f^T_x V'_{x'x'} f_x + V^{'}_{x'} f_{xx}\\
        Q_{xu} & = & l_{xu} + f^T_x V'_{x'x'} f_u + V^{'}_{x'} f_{xu}\\
        Q_{uu} & = & l_{uu} + f^T_u V'_{x'x'} f_u + V^{'}_{x'} f_{uu}.
    \end{IEEEeqnarraybox}
\end{equation}

\subsubsection*{Backward pass}

The optimal control change \(\delta u^*\) is given by minimizing the unconstrained quadratic equation~\eqref{eq:Q_fun_quad_expansion} as
\begin{equation}%
    \label{eq:gains}
    \delta u^* = \underset{u}{\arg\!\min{}} Q(x, u) - u_i = \underbrace{-Q_{uu}^{-1}Q^T_u}_k \underbrace{- Q_{uu}^{-1}Q_{ux}}_K\delta x
\end{equation}
The quadratic approximation of the value function at the current time step in~\eqref{eq:Q_fun_quad_expansion} becomes
\begin{IEEEeqnarray}{rCl}
    \subnumberinglabel{eq:curr_value_function}
    \delta V & = & V(x) - Q(x_i, u_i) = \tfrac{1}{2} Q_u k\\
    V_x & = & Q_x + Q_u K\\
    V_{xx} & = & Q_{xx} + Q_{xu} K,
\end{IEEEeqnarray}
with boundary values \(V^N_x = l_x^N\) and \(V^N_{xx} = l_{xx}^N\).

\subsubsection*{Forward pass}

Once the feedforward and feedback terms \(k_i\) and \(K_i\) for each time step are computed, we perform a forward pass to compute the updated control sequence as
\begin{IEEEeqnarray}{rCl}
    \IEEEyesnumber%
    \IEEEyessubnumber*
    \hat{x}_0 & = & x_0\\
    \hat{u}_i & = & u_i + \delta u^* = u_i + k + K(\hat{x}_i - x_i)\label{eq:feed_control}\\
    \hat{x}_{i+1} & = & f(\hat{x}_i, \hat{u}_i)
\end{IEEEeqnarray}
for \(i \in [0, N-1]\).
In practice, regularization and line search are necessary, as explained in~\cite{tassa2012}.

\subsection{Simulation With Contacts}

We summarize a typical simulation pipeline here in the presence of contacts~\cite{horak2019}.
Contact resolution is usually done in the velocity--impulse level but our DDP is formulated at the acceleration--force level, which will be elaborated later.

The dynamics of a mechanical system are given by
\begin{equation}%
	\label{eq:lagrangian_dynamics}
	M(q)\dot{v} + H(q, v) = S\tau + \sum\nolimits_i J_i^T(q)f_i,
\end{equation}
where \(M\) the mass matrix, \(H\) the vector of nonlinear forces, \(S\) a selection matrix that maps actuated joint torques \(\tau\) to generalized coordinates, while \(J_i\) denotes the Jacobian of the \(i\)-th contact and \(f_i\) the corresponding force.
We simplify notation by dropping explicit dependence on quantities.

In time-stepping approaches, \eg~\cite{todorov2014,horak2019}, \cref{eq:lagrangian_dynamics} is discretized using an Euler approximation to obtain
\begin{equation*}
	M_k\left(v_{k+1} - v_k\right) = h(S \tau_k - H_k) + J^T \lambda_k,
\end{equation*}
where \(h\) is the time step size and \(\lambda_k\) corresponds to the concatenation of the contact impulses at time step \(k\).
These are projected in contact space
\begin{equation*}
	J \left(v_{k+1} - v_k\right) = J M^{-1}_k \left[h (S \tau_k - H_k) + J^T \lambda_k \right],
\end{equation*}
which can also be expressed as
\begin{equation}%
	\label{eq:proj_eom}
	c^+ = A \lambda + b + c^-,
\end{equation}
with \(c^+ = J v_{k+1}\), \(c^- = J v_k\), \(b = h J M^{-1}_k (S \tau_k - H_k)\), and \(A = J M^{-1}_k J^T\).

Different contact models pose different conditions on what constraints accompany \cref{eq:proj_eom}.
In this work, the contact model defined in~\cite{todorov2014} is used because it is convex and analytically invertible.
It penalizes movement in contact space by solving the following QCQP during the forward dynamics
\begin{equation}%
	\label{eq:fd_convex_model}
	\begin{IEEEeqnarraybox}[][c]{r'l}
		\underset{\lambda}{\text{min}} & \tfrac{1}{2} \lambda^T (A + R) \lambda + \lambda^T (b + c^- + c^*)\\
		\text{s.t.} & \lambda_i \in \big \{ \lambda_i \mid \lambda_{n(i)} \geq 0, \norm{\lambda_{t(i)}} \leq \mu_i \lambda_{n(i)}\,\big \} \text{, } \forall i
	\end{IEEEeqnarraybox}
\end{equation}
where \(\lambda_i = \begin{bmatrix} \lambda_{t(i)} & \lambda_{n(i)} \end{bmatrix}^T\) are the tangential and normal components, \(R\) is a positive definite matrix that makes the solution unique and invertible, and \(c^*\) is a Baumgarte stabilization reference.

The inverse dynamics is well-defined and for a diagonal \(R\) we obtain an independent problem per contact
\begin{equation}%
	\label{eq:id_convex_model}
	\begin{IEEEeqnarraybox}[][c]{r'l}
		\underset{\lambda_i}{\text{min}} & \tfrac{1}{2} \lambda_i^T R \lambda_i + \lambda_i^T (c_i^+ + c^*) .\\
		\text{s.t.} & \lambda_i \in \big \{ \lambda_i \mid \lambda_{n(i)} \geq 0, \norm{\lambda_{t(i)}} \leq \mu_i \lambda{n(i)}\,\big \}.
	\end{IEEEeqnarraybox}
\end{equation}

\section{Implicit DDP}%
\label{sec:implicit_DDP}

Our point of departure from the original DPP algorithm is the dynamics in~\eqref{eq:explicit_dynamics}.
Instead of the explicit dynamics, we assume dynamics of the form
\begin{equation}%
	\label{eq:implicit_dynamics}
    g(x', x, u) = 0.
\end{equation}
This will allow us to apply DDP for systems expressed via inverse dynamics, implicit or variational integrators, \etc%
Our focus will be contact dynamics, but we return to this later.

The goal is to compute the derivatives for the quadratic approximation of the \(Q\)-function~\eqref{eq:Q_fun_quad_terms}.
Terms related to the running cost \(l_i\) are trivial and will be omitted.
Thus, we focus on the first and second-order sensitivity of the next step value function.
A treatment of sensitivity analysis in the context of Newton methods can be found in~\cite{zimmermann2019}.

\subsection{First-Order Sensitivity Analysis}

The first-order sensitivity of the value function in~\eqref{eq:Q_fun} is
\begin{equation*}
    V'_x = \tfrac{\partial V'}{\partial x} = \tfrac{\partial V'}{\partial x'} \tfrac{\partial x'}{\partial x} = V'_{x'} \tfrac{\partial x'}{\partial x}.
\end{equation*}
Here, \(V'_x\) is the sensitivity of the next step value function with respect to the current state, while \(V'_{x'}\) is the sensitivity of the next step value function with respect to the next state; connected by the previous equation.
Based on~\eqref{eq:implicit_dynamics} we have
\begin{equation}%
	\label{eq:first_order_sensitivities}
    \tfrac{dg}{dx} = g_{x'} \tfrac{\partial x'}{\partial x} + g_x = 0 \Rightarrow \tfrac{\partial x'}{\partial x} = - g_{x'}^{-1} g_x,
\end{equation}
where it is assumed that for any \(x\) and \(u\), \(x'\) can be computed so that~\eqref{eq:implicit_dynamics} is satisfied.
Combining the previous two equations gives
\begin{equation*}
    V'_x = -V'_{x'} g_{x'}^{-1} g_x.
\end{equation*}
In practice, a faster computation can be achieved using the adjoint method~\cite{strang2007} by computing first the quantity
\begin{equation*}
    s^T = V'_{x'} g_{x'}^{-1} \Rightarrow V'_{x'}{}^T = g_{x'}^T s
\end{equation*}
and then
\begin{IEEEeqnarray}{rCl}%
    \subnumberinglabel{eq:first_order_terms}
    V'_x & = & -s^T g_x.
\end{IEEEeqnarray}
If we confine ourselves in a first-order analysis only this is computationally advantageous~\cite{strang2007}, but the computation of \(\tfrac{\partial x'}{\partial x}\) in~\eqref{eq:first_order_sensitivities} is required for the second-order expansion.
By a similar reasoning, \(V'_u\) is computed as
\begin{IEEEeqnarray}{rCl}
    \IEEEyessubnumber*
    V'_u & = & - s^T g_u,
\end{IEEEeqnarray}
which concludes our first-order analysis.

We now have all the ingredients for the first-order approximation of the \(Q\)-function.
For example, the \(Q_x\) term in~\eqref{eq:Q_fun_quad_terms} is given by
\begin{equation*}
	Q_x = l_x - s^T g_x.
\end{equation*}

\subsection{Second-Order Sensitivity Analysis}

The second-order approximation of the value function is
\begin{equation*}
	V'_{xx} = \tfrac{\partial x'}{\partial x}^T V'_{x'x'} \tfrac{\partial x'}{\partial x} + V'_{x'} \tfrac{\partial^2 x'}{\partial x^2}.
\end{equation*}
The term \(\tfrac{\partial^2 x'}{\partial x^2}\) constitutes a third-order tensor.
We use matrix notation for the contractions but assume that their computation is clear from the context.
It is computed as
\begin{multline*}
	\tfrac{d^2 g}{dx^2} = 0 \Rightarrow \tfrac{\partial^2 x'}{\partial x^2} = g_{x'}^{-1} \left(\tfrac{\partial x'}{\partial x}^T g_{x'x'} \tfrac{\partial x'}{\partial x} + \tfrac{\partial x'}{\partial x}^T g_{x'x}\right.\\
	\left. + \> g_{xx'} \tfrac{\partial x'}{\partial x} + g_{xx}\right).
\end{multline*}
By combining the last two equations we have that
\begin{IEEEeqnarray}{rCl}%
	\subnumberinglabel{eq:tensor_terms}
	V'_{xx} & = & \tfrac{\partial x'}{\partial x}^T V'_{x'x'} \tfrac{\partial x'}{\partial x} - s^T \left(\tfrac{\partial x'}{\partial x}^T g_{x'x'} \tfrac{\partial x'}{\partial x}\right. \nonumber\\
	& & \left. +\> \tfrac{\partial x'}{\partial x}^T g_{x'x} + g_{xx'} \tfrac{\partial x'}{\partial x} + g_{xx}\right).
\end{IEEEeqnarray}

For the remaining two terms in~\eqref{eq:Q_fun_quad_terms}, a similar reasoning can be used to compute them as
\begin{IEEEeqnarray}{rCl}
    \IEEEyessubnumber*
	V'_{xu} & = & \tfrac{\partial x'}{\partial x}^T V'_{x'x'} \tfrac{\partial x'}{\partial u} - s^T \left(\tfrac{\partial x'}{\partial x}^T g_{x'x'} \tfrac{\partial x'}{\partial u} \right. \nonumber\\
	& & \left. +\> \tfrac{\partial x'}{\partial x}^T g_{x'u} + g_{xx'} \tfrac{\partial x'}{\partial u} + g_{xu}\right),\\
	V'_{uu} & = & \tfrac{\partial x'}{\partial u}^T V'_{x'x'} \tfrac{\partial x'}{\partial u} - s^T \left(\tfrac{\partial x'}{\partial u}^T g_{x'x'} \tfrac{\partial x'}{\partial u} \right. \nonumber\\
	& & \left. +\> \tfrac{\partial x'}{\partial u}^T g_{x'u} + g_{ux'} \tfrac{\partial x'}{\partial u} + g_{uu}\right).
\end{IEEEeqnarray}
This concludes the second-order sensitivity analysis.
We can now compute all terms in~\eqref{eq:Q_fun_quad_terms}.
The rest of the DDP algorithm is implemented without changes.

It is worth pointing out that for the explicit dynamics~\eqref{eq:explicit_dynamics} we have that \(g(x', x, u) = f(x,u) - x' = 0\), \(g_{x'} = -I\), and \(g_x = f_x\).
Thus, \(V'_x = V'_{x'} f_x\) as in~\eqref{eq:Q_fun_quad_terms}.
The same verification can be performed for the rest of the quantities.

\subsection{Gauss-Newton Approximation}

Especially for robot models with many degrees of freedom, computing the tensor terms~\eqref{eq:tensor_terms} can be prohibitive expensive.
Fortunately, it is possible to do a Gauss-Newton approximation of the Hessian---equivalent to iLQR---by ignoring them.
Thus, the second-order sensitivity terms of the value function become
\begin{IEEEeqnarray}{rCl}
	\IEEEyesnumber
	\IEEEyessubnumber*
	V'_{xx} & = & \tfrac{\partial x'}{\partial x}^T V'_{x'x'} \tfrac{\partial x'}{\partial x}\\
	V'_{xu} & = & \tfrac{\partial x'}{\partial x}^T V'_{x'x'} \tfrac{\partial x'}{\partial u}\\
	V'_{uu} & = & \tfrac{\partial x'}{\partial u}^T V'_{x'x'} \tfrac{\partial x'}{\partial u}.
\end{IEEEeqnarray}

\section{Acceleration-level Contact Dynamics}%
\label{sec:acceleration_level}

\begin{algorithm}[t]
	\caption{Forward pass with contacts.}%
	\label{al:forward_pass}
	\DontPrintSemicolon
	\KwInput{\(x\), \(k\), \(K\), \(R\), and \(\mu_i\).}
	\KwOutput{\(x'\) and \(f\).}
	Compute \(A+R\) and \(\alpha^- - \alpha^*\) based on \cref{eq:acceleration_space_dynamics,eq:bias_acc}.\\
	Solve \cref{eq:fd_pgs} for the contact forces \(f\).\\
	Solve \cref{eq:implicit_dynamics} together with~\eqref{eq:feed_control} for the next state \(x'\).
\end{algorithm}

\begin{algorithm}[t]
	\caption{Backward pass with contacts.}%
	\label{al:backward_pass}
	\DontPrintSemicolon
	\KwInput{\(x'\), \(x\), \(u\), \(R\), and \(\mu_i\).}
	\KwOutput{\(k\) and \(K\).}
	Compute \(\alpha^+\), \(\alpha^*\), and \(f\) from \cref{eq:acceleration_space_dynamics,eq:bias_acc,eq:contacts_closed_form}.\\
	Compute \cref{eq:implicit_dynamics}.\\
	Differentiate steps 1 and 2 to compute \(g_{x'}\), \(g_x\), \(g_u\), \(g_{x'x'}\), \(g_{x'x}\), \(g_{xx'}\), \(g_{xx}\), \(g_{x'u}\), \(g_{ux'}\), \(g_{xu}\), \(g_{uu}\).\\
	Compute the value function terms in~\eqref{eq:first_order_terms} and~\eqref{eq:tensor_terms}.\\
	Compute the \(Q\)-function terms in \cref{eq:Q_fun_quad_terms}.\\
	Compute the gains \(k\) and \(K\) in \cref{eq:gains}, and the current value function terms in~\eqref{eq:curr_value_function} for the next iteration.
\end{algorithm}

We describe here a contact resolution framework in the acceleration level, rather than the commonly used velocity level.
This way, we avoid the necessary first-order discretization of the dynamics.
Thus, during integration of the state, an arbitrary order integrator can be used.
Other assumptions are not required about the robot's model (such as the assumption about a constant Jacobian in~\eqref{eq:proj_eom} that is inherent in the velocity-impulse formulations), without increasing the computation complexity.
As such, we consider it a superior choice.
It is also the default choice in MuJoCo~\cite{todorov2020}, which is a state-of-the-art robotics simulator.

Starting from the continuous time dynamics~\eqref{eq:lagrangian_dynamics}, we multiply both sides by \(JM^{-1}\) and add \(\dot{J} v\), which gives
\begin{equation}%
	\label{eq:acceleration_space_dynamics}
	\underbrace{J \dot{v} + \dot{J} v}_{\alpha^+} = \underbrace{J M^{-1} J^T}_A f + \underbrace{J M^{-1}(S \tau - H) + \dot{J} v}_{\alpha^-}.
\end{equation}
We can interpret this equation as follows: \(\alpha^-\) is the unconstrained acceleration in contact space in the absence of any contacts, which is corrected by the term \(A f\) to result in the actual acceleration \(\alpha^+\) that satisfies the contact constraints.

As already explained, the contact model that we utilize was proposed in~\cite{todorov2014}.
It computes the necessary contact forces by solving the following convex optimization problem that tries to minimize accelerations in contact space
\begin{equation}%
    \label{eq:fd_pgs}
    \begin{IEEEeqnarraybox}[][c]{r'l}
        \underset{f}{\text{min}} & \tfrac{1}{2} f^T (A + R) f + f^T (\alpha^- - \alpha^*)\\
        \text{s.t.} & f_i \in \big \{ f_i \mid f_{n(i)} \geq 0, \norm{f_{t(i)}}^2 \leq \mu_i^2 f^2_{n(i)}\,\big \},
    \end{IEEEeqnarraybox}
\end{equation}
which is the equivalent to~\eqref{eq:fd_convex_model} for accelerations.

While the bias accelerations \(\alpha^*\) can be in a general Baumgarte stabilization form, a choice that works reasonably good across models is
\begin{equation}%
    \label{eq:bias_acc}
	\alpha^* = \dot{J} v - \tfrac{1}{h^2}\phi(q) - \tfrac{1}{h} J v,
\end{equation}
with \(\phi(q)\) the gap distance, positive when bodies are separate.
The first term is used to cancel the same term in \(\alpha^+\) and \(\alpha^-\) and simplify computations.
The second and third term are obtained by a Taylor expansion of the gap distance function and ignoring third and higher order terms.

In the forward pass, the above optimization problem is solved for the contact forces using a standard Projected Gauss--Siedel solver~\cite{horak2019}.
Though in principle this can be implemented in the backward pass, the computation of the gradients becomes more complicated since we have to differentiate a numeric solution.
Even with automatic differentiation, the quality of the gradients can suffer.
Instead, a diagonal approximation of the system can be assumed and an approximate solution to the contact forces can be computed~\cite{tassa2012}.
The implicit formulation avoids this issue and the exact solution for the contact forces is given in closed form.

\begin{figure}[t]
	\centering
	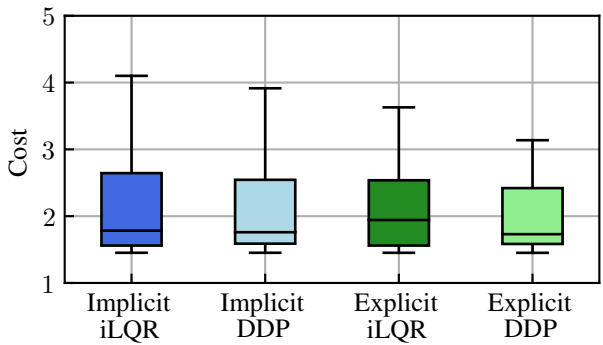
	\vspace{-2mm}
	\caption{Aggregate results for the total trajectory cost of each variant.}%
	\label{fig:cost}
	\vspace{-5mm}
\end{figure}

By utilizing the implicit framework and the invertibility of the model, problem~\eqref{eq:id_convex_model} is expressed in acceleration space
\begin{equation}
    \begin{IEEEeqnarraybox}[][c]{r'l}
        \underset{f}{\text{min}} & \tfrac{1}{2} f^T R f + f^T (\alpha^+ - \alpha^*)\\
        \text{s.t.} & f_i \in \big \{ f_i \mid f_{n(i)} \geq 0, \norm{f_{t(i)}} \leq \mu_i f_{n(i)}\,\big \}.
    \end{IEEEeqnarraybox}
\end{equation}
For the computation of \(\alpha^+\) as given by~\eqref{eq:acceleration_space_dynamics}, the joint acceleration \(\dot{v}\) is required.
In the classical DDP algorithm this is not available, since we only have access to the current state \(q\) and \(v\), and the acceleration is computed after the contact forces.
In the implicit form, since we have additionally available the next state \(x'\), the computation of the acceleration is trivial.
Thus, we can compute each contact force in closed form as
\begin{equation}%
    \label{eq:contacts_closed_form}
	f_i = P_\mu \{-R^{-1} (\alpha^+ - \alpha^*)\}.
\end{equation}
\(P_\mu\) projects contact forces to the cone with coefficient \(\mu\)~\cite{horak2019}.

After the computation of the contact forces, we can enforce the implicit dynamics \cref{eq:implicit_dynamics} either using a forward or inverse dynamics formulation.
Given the available information, the computation of inverse dynamics is cheaper and numerically superior~\cite{hollerbach1980,ferrolho2021}.
Furthermore, this decoupling between forward and backward pass allows us to avoid the rootfinding problem during the forward that would be necessary in a full implicit implementation.
Having to solve the rootfinding problem in the forward pass increases the computation time of the implicit formulation.
We summarize the DDP computations subject to contacts in \cref{al:forward_pass,al:backward_pass}.

\section{Results}%
\label{sec:results}

\begin{figure}[t]
	\centering
	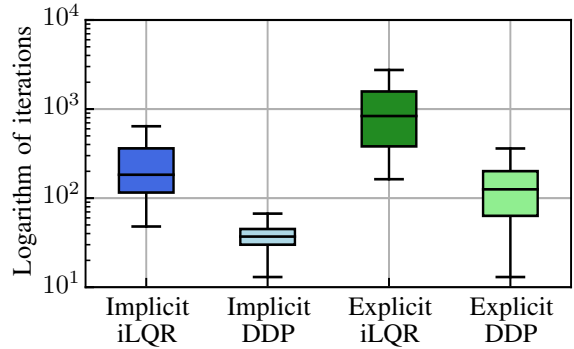
	\vspace{-2mm}
	\caption{Aggregate results for the total number of iterations of each variant.}%
	\label{fig:iters}
	\vspace{-5mm}
\end{figure}

\subsection{Implementation Details}

For the computation of the rigid-body dynamics, the Julia library \texttt{RigidBodyDynamics.jl} is used~\cite{koolen2019}.
Computation of first-, second- and third-order tensor is done using forward-mode automatic differentiation~\cite{revels2016}.
\footnote[0]{The accompanying code is available at \href{https://github.com/ichatzinikolaidis/iDDP}{github.com/ichatzinikolaidis/iDDP} and the video at \href{https://youtu.be/e_TMjmM4NmU}{youtu.be/e\_TMjmM4NmU}.}

We begin by performing multiple comparisons between implicit and explicit DDP formulations for a double pendulum swing-up task.
Next, we present two problems that require multi-contact motion planning: A single leg that is required to a) stand up from the ground, and b) balance from an initial random state.

\subsection{Aggregate Double Pendulum Swing-up}

For the double pendulum swing-up task, we generate \(100\) random trials (that is, with random initial state) and we specify an objective that includes a desired upright posture at the end of a \(T = 5\)s horizon, while penalizing joint torques at intermediate states.
Additionally, joint limits are modelled using unilateral forces at the joints.
Only the unilateral constraint is imposed (forces push the joint away from the limit at violations), while friction is not required.

We compare four variants of methods presented in this work:
\begin{itemize}
	\item Implicit iLQR with backward Euler dynamics.
	\item Implicit DDP with backward Euler dynamics.
	\item Explicit iLQR with forward Euler dynamics.
	\item Explicit DDP with forward Euler dynamics.
\end{itemize}
For every random initialization, the four variants are executed until convergence (or until an upper iteration limit is reached) and the number of iterations and total cost of the trajectory is logged.
Aggregate box plot results for the cost and the number of iterations are shown in~\cref{fig:cost,fig:iters}, respectively.

From the comparison, the implicit formulations result in considerably less iterations than the explicit counterpart.
Both median, minimum and maximum values, and the rest of the statistical properties in~\cref{fig:iters} are improved with an implicit formulation regarding the number of iterations.
As expected, the trade-off for this is the larger in general cost of the resulting trajectory in~\cref{fig:cost}.
This can be partially explained from the fact that since the explicit formulations perform on average more iterations, they are capable to fine-tune the resulting trajectory more.
But given the considerable less iterations for the implicit formulations, this aspect is more important in terms of the overall performance.

A possible reason behind this is the integrator's properties.
Implicit Euler is an A-stable method suitable even for stiff systems.
As such, it usually exhibits energy decrease---instead of the common increase in explicit methods---that makes the whole formulation more stable.

\subsection{Single Double Pendulum Swing-up}

\begin{figure}[t]
	\centering
	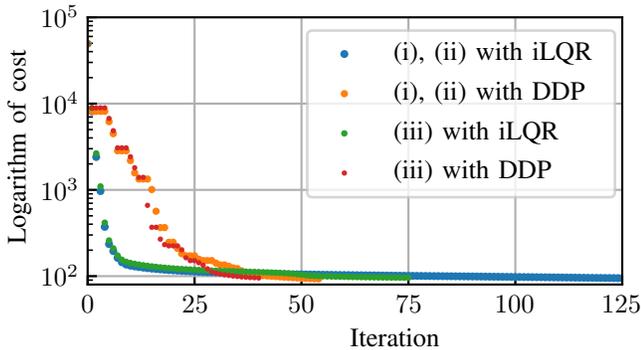
	\vspace{-2mm}
	\caption{Cost per iteration for the different formulations.}%
	\label{fig:cost_per_iter}
	\vspace{-5mm}
\end{figure}

\subsubsection{Cost per iteration and timings}

We evaluate the cost per iteration for one double pendulum swing-up and compare 6 different formulations (each with a DDP and iLQR variant):
\begin{enumerate}[label=(\roman*),align=left,leftmargin=*,labelsep=-4pt]
    \item Forward Euler dynamics in the forward and backward passes.
    \item Forward Euler dynamics in the forward pass, and forward Euler inverse dynamics in the backward pass.
    \item Backward Euler dynamics in the forward and backward passes.
\end{enumerate}
We use the same cost function and initialize at the stable equilibrium, while the total duration of the motion is \(5\)s, with a time step of \(10\)ms, and \(500\) steps.
The results are shown in \cref{fig:cost_per_iter}.
Formulation (i) corresponds to a classical iLQR/DDP with explicit dynamics.
Formulation (ii) is enabled by the presented framework.
The computation of the Jacobian and tensor terms is based on the automatic differentiation of the inverse dynamics.
Since (ii) is equivalent to (i), the solutions by the two approaches are exactly the same and are plotted together in \cref{fig:cost_per_iter}.
Differences are found for the computation time, as reported next.
Formulation (iii) is implicit in both passes and enabled again by the presented framework.

In terms of computation, formulations (i) and (ii) with iLQR require \(126\) iterations, while with DDP require \(55\) iterations.
In terms of timings, the mean time of each iteration for (i) with iLQR is \(5.87\)ms and with DDP \(29.91\)ms.
For (ii) with iLQR is \(5.03\)ms and with DDP \(28.49\)ms.
While the differences are not significant for such a low-dimensional model, these can become starker for robot models with larger degrees of freedom.
For (iii), \(75\) iterations for iLQR and \(40\) iterations with DDP.
The mean computation time of each iterations with iLQR is \(7.19\)ms and with DDP \(72.22\)ms.
The increased computation is due to the solution of a nonlinear system of equations in the forward pass.

\subsubsection{Effect of time step size}

We focus now on the effect of the time step size to the solution of the problem.
We solve the same problem as before for multiple time step selections and report the number of iterations required until convergence.
Since formulations (i) and (ii) are equivalent, we focus the comparison on (i) and (iii).
We solve them using iLQR but similar conclusions could be drawn if DDP was used.

The results are shown on \cref{tab:time_step_effect}.
For small time steps, the two formulations are essentially equivalent and, thus, require the same number of iterations.
As the time step increases, the influence of the integrator's damping in (iii) becomes more apparent.
This results in a desirable decrease to the number of iterations for convergence.
The motions are included again in the accompanying video.
For larger time steps, the accuracy of both first-order integrators worsens significantly.

\begin{table}[t]
    \centering
    \caption{Effect of time step on No. of iterations until convergence}%
    \label{tab:time_step_effect}
    \begin{tabular}{c|ccc}
        \toprule
        Time step  & \(10^{-4}\) & \(10^{-3}\) & \(10^{-2}\)\\
        \midrule
        (i) / (ii) & \(56\)      & \(68\)      & \(126\)\\
        (iii)      & \(56\)      & \(66\)      & \(75\)\\
        \bottomrule
    \end{tabular}
	\vspace{-5mm}
\end{table}

\subsection{Multi-contact Stand-up}

\begin{figure*}[t]
	\centering
	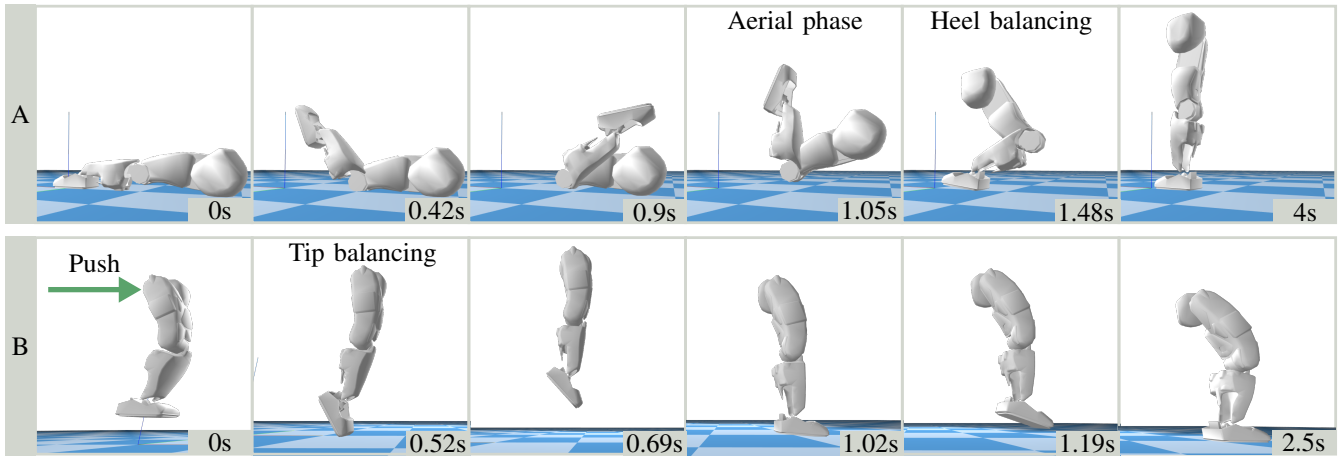
	\caption{Time-lapse snapshots of contact-rich motions:
	(A) Standing-up from ground by a dynamic manoeuvre using large momentum with aerial phases.
	(B) Robust balancing by toe tipping and jumping that withstands an external push.}%
	\label{fig:combination}
	\vspace{-5mm}
\end{figure*}

Next, we consider a planar 3 degree-of-freedom single leg of a humanoid robot and the task now is to stand-up upright from the ground.
The model can make multiple contacts with the terrain using all the bodies of its structure, but self-collisions are inactive.
We pre-define a number of possible contact points but we do not prescribe the contact activation pattern.
Adding a contact detection mechanism and avoiding the pre-specification of contacts is another possibility, as typically done in simulation engines.

The cost function of the problem is defined as
\begin{equation*}
	J = w_{q_f} ||q_f - q_\text{g}||^2 + w_{v_f} ||v_f||^2 + \sum\nolimits_i (w_\tau ||\tau_i||^2 + w_v ||v_i||^2).
\end{equation*}
A penalization of the velocity and joint torque is applied throughout the trajectory, while a goal state is defined in the final cost term.
The motion duration is \(T=4\)s with a time step of \(10\)ms; this is a relatively large time step for contacts, but our aim here is to output an approximate contact-rich motion plan.
Given this plan as input, it is possible to post-process it to increase the quality.

The friction coefficient is selected as \(\mu=0.7\).
Parameter \(R\) is initialized with a value of \(1\) for all components.
While in principle it can take arbitrary values, we can test the validity from a numerical viewpoint as follows~\cite{todorov2020}: We run the forward and backward pass separately and compare the computed forces.
The two solutions should match according to the desired numerical precision.

% Snapshots of the computed motion are shown at the top row of \cref{fig:combination}.
The main difficulty is that the problem exhibits a number of contact possibilities.
Thus, mode enumeration can be very challenging.
Notice also how delicate heel balance emerges while reaching the upright configuration.
Our trajectory optimization framework is capable to output a locally optimal motion plan.
Even though a zero torque initial solution is used here, its quality greatly affects the quality of the computed motion.
Finally, by changing the terms in the cost function, it is possible to obtain different solutions, \eg~more conservative but with higher torque cost.

The resulting motion can be found in the accompanying video.
There is an initial explosive and dynamic motion at about \(1\)s.
Such a motion would be in practice difficult to track.
Yet being able to compute such a complex motion from high-level input only demonstrates the power of DDP-based approaches.
There are a couple of ways to mitigate that: an obvious approach is to increase the torque, position, and velocity penalization accordingly.
Another option is to include terms that penalize the rate of the commanded torques.
Finally, a more principled approach is to penalize high frequency components of the signals involved~\cite{grandia2019}.

\subsection{Multi-contact Balancing}

Using the same model as before, the state now is randomly initialized in the air.
The task is to keep the initial posture with zero velocity, \ie~to balance.
In contrast to the previous case, this problem is formulated in a receding horizon fashion.
A fixed number of \(15\) iterations for DDP is pre-specified; this makes real-time iterations of the algorithm possible.
The horizon length is \(T=0.5\)s, with the simulation running at \(200\)Hz, while our framework runs at \(20\)Hz. 
The structure of the cost function remains the same as before, albeit the weight regarding the final velocity is increased to bias more towards a static final configuration.

The balancing motion is shown at the bottom of \cref{fig:combination}.
The computed motion naturally performs a series of jumps to dissipate kinetic energy and come to a complete stop.
The underactuated foot tilting emerged as the outcome of optimization without programming explicit controllers as in \cite{li2017}.
Compared to the case in the previous section, the receding horizon formulation is capable of producing better motions in general.
This is because the constant updates allows it to escape iterations with very small cost decrease, which can be common in the fixed horizon optimization of the previous case.
If a bad initialization is specified or the horizon and frequency are not chosen properly, the receding horizon formulation can be trapped too.
The selection of these parameters depends on the desired task and initial state.

\begin{figure}[t]
    \centering
	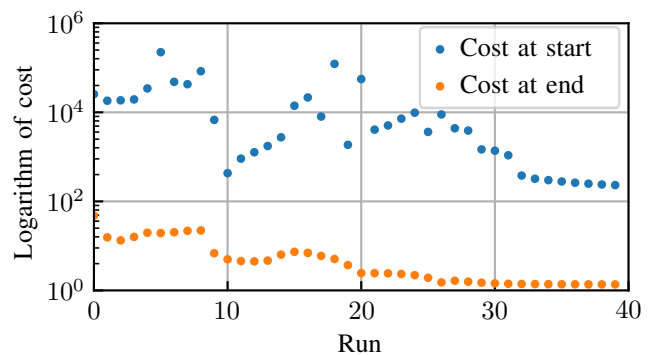
	\vspace{-2mm}
    \caption{Trajectory cost at each run of the receding horizon formulation.}%
	\label{fig:val_iters}
	\vspace{-5mm}
\end{figure}

Finally, a semi-log plot of the total trajectory cost at the beginning and at the end of each DDP step is shown in~\cref{fig:val_iters}.
We notice that in about \(20\) runs a successful balancing motion is computed.
Afterwards, each run rapidly converges to this motion.
The reason why the cost is increased at the beginning of each run is because the horizon moves; the predicted trajectory for the new segment at the end of the previous horizon is that the robot will essentially fall, which incurs a large cost.
Additionally, during the initial runs, the motion is highly unstable and a suitable balancing motion is not discovered yet.
Thus, the total trajectory cost varies greatly between consecutive runs.

\section{Conclusion}%
\label{sec:conclusion}

This work presents an extension to DDP that is able to handle implicit dynamical systems with particular focus on through-contact motion planning.
This allows extending the original DDP to a larger class of dynamics models, \eg~such as models based on inverse dynamics.
We described how to use the implicit formulation for accurate contact resolution in the DDP framework without requiring approximations of contact dynamics.
The proposed method is exact and straightforward to implement, utilizing a closed-form solution for quality gradient computations.
Further, we demonstrated properties of the approach in a number of cases: comparisons of implicit and explicit dynamics representations for a double pendulum, and two case studies for a single leg model that required challenging multi-contact motion plans.

While the original DDP provides both feedforward and feedback gains that guarantee a level of robustness against small perturbations, we noticed that the computed motion plans can fail if the conditions of the problem change slightly.
Though one can introduce robustness as part of the trajectory optimization modelling, we believe that running the whole framework in a receding horizon fashion is more appropriate and promising.
Thus, the motion plans should be updated online to withstand unexpected perturbations.

It is worth noting that DDP simulates the dynamics of the system and activates a contact point if it finds one.
Thus, contacts are taken into account according to the system's natural dynamics~\cite{posa2014}, which may lead to characteristically abrupt motions~\cite{tassa2012}.
Being a shooting method for unconstrained systems, DDP is limited in terms of active search for potential contacts.
Further improvements can be made by combinatorial planning and exploration, where transcription-based methods demonstrated better capabilities and flexibility~\cite{chatzinikolaidis2020,patel2019}, although requiring additional and non-negligible computation cost in practice.

%%%%%%%%%%%%%%%%%%%%%%%%%%%%%%%%%%%%%%%%%%%%%%%%%%%%%%%%%%%%%%%%%%%%%%%%%%%%%%%%

\printbibliography

\end{document}